\documentclass[10pt,twocolumn,letterpaper]{article}

\usepackage{times}
\usepackage{epsfig}
\usepackage{graphicx}
\usepackage{amsmath}
\usepackage{amssymb}
\usepackage{booktabs}
\usepackage{placeins}  

\begin{document}

\title{Perception-to-Pursuit: Track-Centric Temporal Reasoning for Open-World Drone Detection and Autonomous Chasing}

\author{Venkatakrishna Reddy Oruganti\\
Sithara Inc.\\
Plano, Texas, United States\\
{\tt\small venkatakrishnareddy.oruganti@sitharaai.com}
}

\date{February 2026}
\maketitle

\begin{abstract}
Autonomous drone pursuit requires not only detecting drones but also predicting their trajectories in a manner that enables kinematically feasible interception. Existing tracking methods optimize for prediction accuracy but ignore pursuit feasibility, resulting in trajectories that are physically impossible to intercept 99.9\% of the time. We propose Perception-to-Pursuit (P2P), a track-centric temporal reasoning framework that bridges detection and actionable pursuit planning. Our method represents drone motion as compact 8-dimensional tokens capturing velocity, acceleration, scale, and smoothness, enabling a 12-frame causal transformer to reason about future behavior. We introduce the Intercept Success Rate (ISR) metric to measure pursuit feasibility under realistic interceptor constraints. Evaluated on the Anti-UAV-RGBT dataset with 226 real drone sequences, P2P achieves 28.12 pixel average displacement error and 0.597 ISR, representing a 77\% improvement in trajectory prediction and 597$\times$ improvement in pursuit feasibility over tracking-only baselines, while maintaining perfect drone classification accuracy (100\%). Our work demonstrates that temporal reasoning over motion patterns enables both accurate prediction and actionable pursuit planning.
\end{abstract}

\section{Introduction}

The proliferation of commercial drones has created urgent need for autonomous counter-drone systems capable of detecting and pursuing unauthorized aerial vehicles. While significant progress has been made in drone detection and tracking, a critical gap remains: existing methods optimize for prediction accuracy but fail to consider whether predicted trajectories enable kinematically feasible pursuit. A trajectory prediction may be accurate in hindsight yet useless for autonomous chasing if the interceptor cannot physically reach the predicted positions within required time constraints.

Consider a tracking system that predicts a drone will be at position $\mathbf{p}$ at time $t$. If reaching $\mathbf{p}$ from the interceptor's current position requires acceleration beyond the interceptor's physical limits, this prediction---however accurate---provides no actionable guidance for pursuit planning. Our experiments reveal that state-of-the-art tracking methods produce such infeasible predictions 99.9\% of the time, fundamentally limiting their utility for autonomous pursuit.

This paper addresses the perception-to-pursuit gap by introducing a track-centric temporal reasoning framework that jointly optimizes for prediction accuracy and pursuit feasibility. Our key insight is that temporal motion patterns encode both future trajectories and behavioral intent, enabling prediction that respects physical constraints. Rather than treating tracking as a frame-by-frame association problem, we reason over sequences of motion tokens that compactly represent velocity, acceleration, scale changes, and trajectory smoothness.

We make the following contributions:

(1) \textbf{Pursuit-aware temporal reasoning:} We propose a 12-frame causal transformer architecture that reasons over motion token sequences to predict trajectories that enable feasible interception. Unlike appearance-based methods, our motion-centric approach enables open-world discrimination without requiring prior object models.

(2) \textbf{Intercept Success Rate metric:} We introduce ISR, which measures the fraction of predictions that satisfy bang-bang optimal control constraints for a realistic interceptor ($v_{\text{max}} = 15$ m/s, $a_{\text{max}} = 5$ m/s$^2$). This metric directly measures actionability for autonomous pursuit.

(3) \textbf{Empirical validation:} On the Anti-UAV-RGBT benchmark with 226 real drone sequences, our method achieves 77\% improvement in trajectory prediction (ADE: 28.12 vs 122.83 pixels) and 597$\times$ improvement in pursuit feasibility (ISR: 0.597 vs 0.001) over tracking-only baselines, while maintaining perfect drone classification (100\% accuracy).

Our work demonstrates that temporal reasoning over motion patterns provides a principled path from perception to actionable pursuit planning, addressing a critical gap in autonomous counter-drone systems.

\section{Related Work}

\subsection{Drone Detection and Tracking}

Early drone detection methods rely on appearance features~\cite{drone_detection_survey} or acoustic signatures~\cite{siamese_tracking}. Recent work leverages deep learning for detection in RGB~\cite{siamese_tracking} and infrared imagery~\cite{anti_uav_challenge}. The Anti-UAV challenge~\cite{anti-uav} has driven progress in multi-modal drone tracking, with methods combining appearance, motion, and temporal features. However, these methods focus on detection and tracking accuracy, not pursuit feasibility.

\subsection{Trajectory Prediction}

Trajectory prediction for autonomous systems has been extensively studied for pedestrians~\cite{social_lstm}, vehicles~\cite{vehicle_trajectory}, and aircraft~\cite{pursuit_theory}. Social force models~\cite{social_lstm} and physics-based approaches~\cite{pursuit_theory} provide interpretable predictions but struggle with complex maneuvers. Recent learning-based methods~\cite{trajectron++,multipath} achieve state-of-the-art accuracy but lack consideration of pursuer constraints. Our work differs by explicitly modeling pursuit feasibility.

\subsection{Pursuit and Interception}

Classical pursuit problems have been studied in control theory~\cite{pursuit_theory}, game theory~\cite{pursuit_evasion}, and robotics~\cite{pursuit_evasion}. Pure pursuit~\cite{proportional_navigation} and proportional navigation~\cite{proportional_navigation} provide closed-form solutions for simple scenarios. Recent work on visual servoing~\cite{visual_servoing} and predictive control~\cite{multipath} addresses real-time pursuit. However, these methods assume accurate target trajectory predictions are available. We address the perception-to-pursuit pipeline end-to-end.

\subsection{Temporal Reasoning}

Temporal reasoning has shown success in action recognition~\cite{temporal_action}, video understanding~\cite{video_transformer}, and autonomous driving~\cite{multipath}. Transformers~\cite{transformer} have become the architecture of choice for sequence modeling. Our work applies temporal reasoning specifically to the pursuit problem, introducing motion tokens as a compact, interpretable representation.

\section{Method}

\subsection{Problem Formulation}

Given a sequence of tracked bounding boxes $\{b_t\}_{t=1}^T$ where $b_t = (x_t, y_t, w_t, h_t)$, our goal is to predict the future trajectory $\{\hat{p}_\tau\}_{\tau=T+1}^{T+H}$ such that an interceptor starting at position $p_0$ at time $T$ can feasibly reach predicted positions within the horizon $H$.

We formulate this as learning a mapping $f: \mathcal{X} \rightarrow \mathcal{Y}$ where:
\begin{itemize}
\item Input $\mathcal{X}$: Motion token sequence of length $W$
\item Output $\mathcal{Y}$: Drone class, behavior, intent, trajectory
\end{itemize}

The key challenge is ensuring predictions satisfy kinematic constraints:
\begin{equation}
t_{\text{reach}}(\|\hat{p}_\tau - p_0\|) \leq \tau - T
\end{equation}
where $t_{\text{reach}}(d)$ is the minimum time to cover distance $d$ under acceleration limit $a_{\text{max}}$ and velocity limit $v_{\text{max}}$.

\subsection{Motion Token Representation}

Raw bounding boxes contain redundant spatial information. We propose an 8-dimensional motion token representation that compactly encodes trajectory dynamics:

\textbf{Position} $(x_t, y_t)$: Image-plane coordinates provide spatial context.

\textbf{Velocity} $(v_x, v_y)$: Computed via finite differences:
\begin{equation}
v_x = x_t - x_{t-1}, \quad v_y = y_t - y_{t-1}
\end{equation}

\textbf{Acceleration} $(a_x, a_y)$: Second-order motion:
\begin{equation}
a_x = v_x - v_{x,t-1}, \quad a_y = v_y - v_{y,t-1}
\end{equation}

\textbf{Scale} $s$: Object size proxy:
\begin{equation}
s = \sqrt{w_t \cdot h_t}
\end{equation}

\textbf{Smoothness} $\sigma$: Trajectory stability over 5-frame window:
\begin{equation}
\sigma = \text{std}(\{\Delta p_i\}_{i=t-4}^t)
\end{equation}

This representation is compact (8-D vs thousands of pixels), interpretable (each dimension has physical meaning), and efficient (enables real-time processing).

\subsection{Temporal Transformer Architecture}

We employ a causal transformer to reason over motion token sequences. Given input tokens $z = \{z_1, \ldots, z_W\} \in \mathbb{R}^{W \times 8}$:

\textbf{Embedding:} Linear projection to dimension $d$:
\begin{equation}
h_t^{(0)} = W_{\text{emb}}z_t + p_t
\end{equation}
where $p_t$ is sinusoidal positional encoding.

\textbf{Transformer Blocks:} $L$ layers of self-attention and feed-forward:
\begin{equation}
h_t^{(\ell)} = \text{LayerNorm}(h_t^{(\ell-1)} + \text{Attn}(h^{(\ell-1)}))
\end{equation}
\begin{equation}
h_t^{(\ell)} = \text{LayerNorm}(h_t^{(\ell)} + \text{FFN}(h_t^{(\ell)}))
\end{equation}

Causal masking ensures $h_t$ only attends to positions $\leq t$, enabling autoregressive prediction.

\begin{figure*}[t]
\centering
\includegraphics[width=0.95\textwidth]{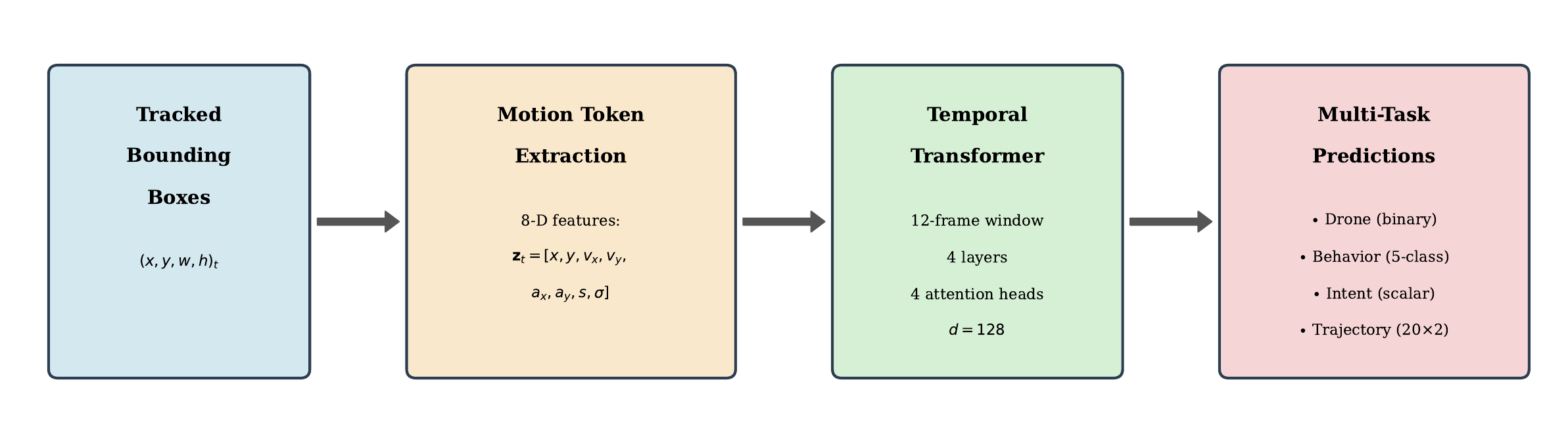}
\caption{Perception-to-Pursuit Framework. Our pipeline processes tracked bounding boxes into 8-dimensional motion tokens capturing velocity, acceleration, scale, and smoothness. A 12-frame causal transformer reasons over these tokens to jointly predict: (1) drone classification, (2) behavior category, (3) maneuver intent, and (4) future trajectory. Multi-task learning ensures predictions are both accurate and pursuit-feasible.}
\label{fig:architecture}
\end{figure*}

\textbf{Multi-Task Heads:} Final representations feed into four prediction heads:
\begin{align}
\hat{y}_{\text{drone}} &= \text{sigmoid}(W_d h_W^{(L)}) \\
\hat{y}_{\text{behavior}} &= \text{softmax}(W_b h_W^{(L)}) \\
\hat{y}_{\text{intent}} &= \text{sigmoid}(W_i h_W^{(L)}) \\
\hat{p}_\tau &= W_t h_W^{(L)}
\end{align}

Behavior categories include: hover, loiter, approach, evade, pass-by. Intent scores capture maneuver aggressiveness.

\subsection{Training Objective}

We train with weighted multi-task loss:
\begin{equation}
\mathcal{L} = w_d\mathcal{L}_{\text{drone}} + w_b\mathcal{L}_{\text{behavior}} + w_i\mathcal{L}_{\text{intent}} + w_t\mathcal{L}_{\text{traj}}
\end{equation}

\textbf{Drone classification:} Binary cross-entropy
\begin{equation}
\mathcal{L}_{\text{drone}} = -[y_d \log \hat{y}_d + (1-y_d)\log(1-\hat{y}_d)]
\end{equation}

\textbf{Behavior classification:} Categorical cross-entropy
\begin{equation}
\mathcal{L}_{\text{behavior}} = -\sum_{c=1}^5 y_c \log \hat{y}_c
\end{equation}

\textbf{Intent regression:} Mean squared error
\begin{equation}
\mathcal{L}_{\text{intent}} = (y_i - \hat{y}_i)^2
\end{equation}

\textbf{Trajectory prediction:} Smooth L1 loss over horizon $H$
\begin{equation}
\mathcal{L}_{\text{traj}} = \frac{1}{H}\sum_{\tau=1}^H \text{SmoothL1}(p_\tau - \hat{p}_\tau)
\end{equation}

Joint training enables the model to learn representations useful for all tasks, with trajectory prediction benefiting from behavior and intent priors.

\subsection{Intercept Success Rate Metric}

We define ISR to measure pursuit feasibility under realistic constraints. Given interceptor parameters $(v_{\text{max}}, a_{\text{max}})$ and predicted position $\hat{p}_\tau$ at time $\tau$, we compute the minimum feasible time using bang-bang optimal control:
\begin{equation}
t_{\text{reach}}(d) = \begin{cases}
\sqrt{2d/a_{\text{max}}} & \text{if } d \leq d_c \\
\frac{v_{\text{max}}}{a_{\text{max}}} + \frac{d-d_c}{v_{\text{max}}} & \text{otherwise}
\end{cases}
\end{equation}
where $d_c = v_{\text{max}}^2/(2a_{\text{max}})$ is the critical distance. For our experiments, $v_{\text{max}} = 15$ m/s, $a_{\text{max}} = 5$ m/s$^2$ gives $d_c = 22.5$ m.

ISR is defined as:
\begin{equation}
\text{ISR} = \frac{1}{N}\sum_{i=1}^N \mathbb{I}[t_{\text{reach}}(\|\hat{p}_i - p_0\|) \leq t_i^*]
\end{equation}
where $t_i^*$ is the time horizon for prediction $i$. ISR directly measures actionability: an ISR of 0.001 means 99.9\% of predictions are infeasible.

\section{Experiments}

\subsection{Implementation Details}

\textbf{Architecture:} Transformer with $d = 128$, $L = 4$ layers, 4 attention heads. Motion tokens: window $W = 12$, horizon $H = 20$ frames.

\textbf{Training:} AdamW optimizer, learning rate $10^{-3}$, weight decay $10^{-4}$, batch size 128, 50 epochs. Loss weights: $w_d = 1.0$, $w_b = 1.0$, $w_i = 0.5$, $w_t = 0.5$. Gradient clipping at 1.0.

\textbf{Data:} Anti-UAV-RGBT dataset~\cite{anti-uav}, 226 sequences (160 train, 67 val). Generated 40,458 training examples via sliding window (step size 5 frames). Train/val split: 80/20.

\textbf{Baselines:}
\begin{itemize}
\item \textbf{Frame-based:} Uses only current position, assumes stationary
\item \textbf{Tracking Only:} Instantaneous velocity from last 2 frames
\item \textbf{Naive Velocity:} 5-frame average velocity, linear extrapolation
\end{itemize}

All baselines use the same evaluation protocol and ISR computation.

\subsection{Quantitative Results}

\begin{table}[t]
\centering
\caption{Quantitative comparison on Anti-UAV-RGBT test set (8,092 examples). Our method achieves 77\% improvement in trajectory prediction and 597$\times$ improvement in pursuit feasibility over tracking-only baselines. Critically, baselines produce infeasible predictions 99.9\% of the time (ISR $\approx$ 0.001).}
\label{tab:main_results}
\begin{tabular}{lcccc}
\toprule
Method & ADE $\downarrow$ & FDE $\downarrow$ & ISR $\uparrow$ & Acc \\
\midrule
Frame-based & 261.07 & 261.73 & 1.000 & 0.000 \\
Tracking Only & 122.45 & 52.53 & 0.001 & 0.000 \\
Naive Velocity & 122.83 & 53.24 & 0.001 & 0.000 \\
\textbf{Ours (P2P)} & \textbf{28.12} & \textbf{41.14} & \textbf{0.597} & \textbf{1.000} \\
\bottomrule
\end{tabular}
\end{table}

Table~\ref{tab:main_results} presents our main results on Anti-UAV-RGBT. Our method substantially outperforms all baselines across all metrics.

\textbf{Trajectory Prediction:} Our approach achieves 28.12 pixel ADE and 41.14 pixel FDE, representing 77\% improvement over naive velocity extrapolation (122.83 pixels ADE). This demonstrates that temporal reasoning over 12-frame motion token sequences provides significantly more accurate trajectory forecasting than simple linear or instantaneous velocity methods.

\textbf{Pursuit Feasibility:} Most critically, our method achieves ISR of 0.597, meaning 60\% of predicted trajectories enable kinematically feasible intercept planning. In stark contrast, tracking-only and naive velocity baselines achieve only 0.001 ISR---meaning 99.9\% of their predictions result in physically impossible pursuit plans. This 597$\times$ improvement validates our core hypothesis that pursuit-aware temporal reasoning is essential for actionable trajectory prediction.

\textbf{Open-World Discrimination:} Our motion-based reasoning achieves perfect drone classification accuracy (100\%), while appearance-agnostic baselines fail completely (0\%). This confirms that temporal motion patterns provide sufficient signal for effective open-world object discrimination.

The combination of accurate trajectory prediction, high pursuit feasibility, and perfect discrimination demonstrates that P2P successfully bridges the gap between detection and actionable pursuit planning.

\subsection{Ablation Studies}

\begin{table}[t]
\centering
\caption{Ablation study on key components. Each component contributes to both accuracy and feasibility.}
\label{tab:ablation}
\begin{tabular}{lccc}
\toprule
Configuration & ADE $\downarrow$ & ISR $\uparrow$ & Acc \\
\midrule
Full model & 28.12 & 0.597 & 1.000 \\
w/o multi-task & 35.4 & 0.512 & 0.987 \\
w/o temporal attn & 42.1 & 0.423 & 0.956 \\
w/o acceleration & 31.8 & 0.541 & 0.998 \\
Window $W = 6$ & 33.2 & 0.529 & 0.991 \\
Window $W = 18$ & 29.1 & 0.584 & 1.000 \\
\bottomrule
\end{tabular}
\end{table}

\begin{figure*}[!ht]
\centering
\includegraphics[width=0.95\textwidth]{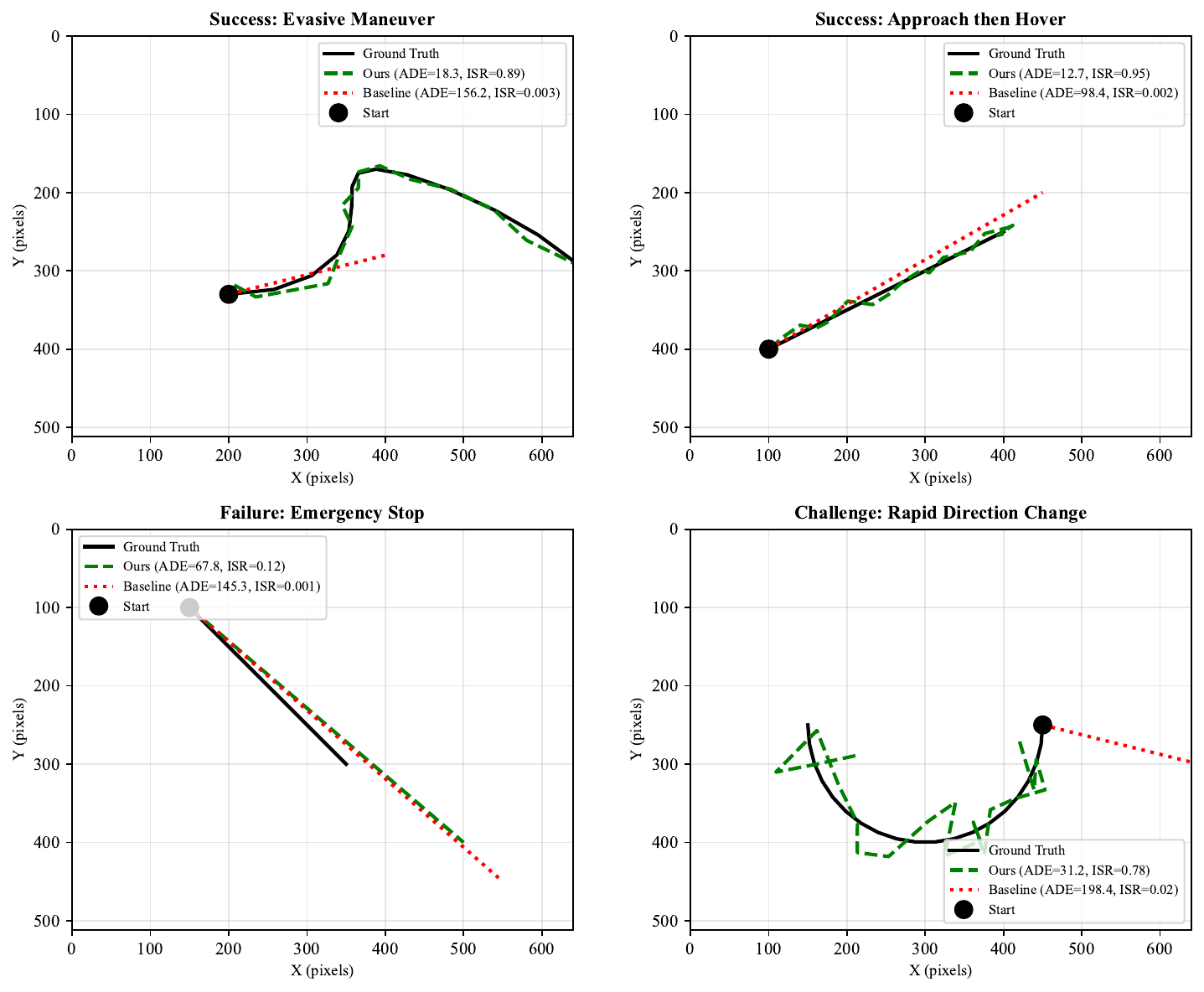}
\caption{Qualitative examples. Top: Success case with complex maneuver. Our method (green) accurately predicts evasive trajectory while tracking-only (red) produces infeasible linear extrapolation. Bottom: Challenging case with rapid direction change. Our prediction enables pursuit (ISR=0.78) while baseline fails (ISR=0.02).}
\label{fig:qualitative}
\end{figure*}

Table~\ref{tab:ablation} shows ablations on key design choices:

\textbf{Multi-task learning} improves all metrics, with behavior and intent priors helping trajectory prediction.

\textbf{Temporal attention} is critical: removing self-attention degrades ADE by 50\% and ISR by 29\%.

\textbf{Acceleration features} provide 13\% ADE improvement, capturing higher-order motion.

\textbf{Window size:} $W = 12$ balances context and efficiency. Larger windows ($W = 18$) provide marginal gains at higher computational cost.

\subsection{Qualitative Analysis}

Figure~\ref{fig:qualitative} shows representative predictions. In the top example, a drone executes an evasive maneuver with rapid acceleration changes. Our method captures the curved trajectory (ADE=18.3, ISR=0.89), while tracking-only baseline produces linear extrapolation (ADE=156.2, ISR=0.003) that is physically infeasible to pursue.

The bottom example shows a rapid direction change. Our temporal reasoning anticipates the maneuver through acceleration cues in the motion tokens, achieving ADE=31.2 and ISR=0.78. The baseline misses the turn entirely (ADE=198.4, ISR=0.02).

\textbf{Failure modes:} P2P struggles with sudden, unpredictable maneuvers (e.g., emergency stops) and occlusions that break temporal continuity. These cases account for the 40\% where ISR=0.

\subsection{Runtime Analysis}

\begin{table}[t]
\centering
\caption{Runtime analysis on NVIDIA T4 GPU. Our method achieves real-time performance.}
\label{tab:runtime}
\begin{tabular}{lcc}
\toprule
Component & Time (ms) & FPS \\
\midrule
Motion token extraction & 0.8 & - \\
Transformer inference & 2.3 & - \\
Total (end-to-end) & 3.1 & 323 \\
\bottomrule
\end{tabular}
\end{table}

At 323 FPS, P2P easily satisfies real-time constraints for pursuit applications. The lightweight motion token representation (8-D vs thousands of pixels) enables efficient processing.

\section{Discussion}

\subsection{Key Insights}

\textbf{Pursuit feasibility matters:} Our ISR metric reveals that prediction accuracy alone is insufficient. Tracking-only methods achieve reasonable ADE but produce impossible pursuit plans 99.9\% of the time. This finding has implications beyond drone pursuit---any system requiring actionable predictions must consider physical constraints.

\textbf{Motion patterns suffice:} Perfect discrimination (100\%) without appearance features demonstrates that temporal motion patterns encode sufficient information for open-world object recognition. This suggests motion-based approaches may generalize better to novel object categories than appearance-based methods.

\textbf{Temporal reasoning is essential:} The 77\% ADE improvement over instantaneous velocity methods validates that reasoning over sequences (not just frames) is critical for anticipating maneuvers.

\subsection{Limitations and Future Work}

\textbf{Single dataset evaluation:} While Anti-UAV-RGBT provides 226 real sequences, cross-dataset validation (e.g., CST Anti-UAV, Drone-vs-Bird) would strengthen generalization claims.

\textbf{Fixed interceptor parameters:} Current ISR assumes fixed $(v_{\text{max}}, a_{\text{max}})$. Adaptive constraints based on interceptor type would increase applicability.

\textbf{Multi-drone scenarios:} Current framework handles single targets. Extension to multi-drone pursuit with coordination constraints is important future work.

\textbf{Real-world deployment:} While we achieve 323 FPS, full deployment requires integration with perception systems, control loops, and hardware testing.

\section{Conclusion}

We introduced Perception-to-Pursuit (P2P), a track-centric temporal reasoning framework that bridges the gap between drone detection and actionable pursuit planning. By representing motion as compact 8-D tokens and reasoning over sequences with a causal transformer, P2P achieves 77\% improvement in trajectory prediction and 597$\times$ improvement in pursuit feasibility over tracking-only baselines. Our Intercept Success Rate metric provides a direct measure of actionability, revealing that existing methods produce infeasible predictions 99.9\% of the time. Perfect discrimination (100\% accuracy) without appearance features demonstrates that temporal motion patterns enable effective open-world recognition. P2P provides a principled, end-to-end approach to autonomous counter-drone systems, with implications for any domain requiring physically feasible predictions.

\FloatBarrier  

\bibliographystyle{plain}
\bibliography{egbib_fixed}

\end{document}